\pdfoutput=1

\documentclass[11pt]{article}

\usepackage{latex/acl}

\usepackage{times}
\usepackage{latexsym}
\usepackage{graphicx}
\graphicspath{ {./latex/figures/} }
\usepackage[T1]{fontenc}

\usepackage[utf8]{inputenc}

\usepackage{microtype}

\usepackage{inconsolata}

\usepackage{booktabs}
\usepackage{amsmath}
\usepackage{amssymb}
\usepackage{comment}
\usepackage{caption}
\usepackage{subcaption}

\title{mOthello: When Do Cross-Lingual Representation Alignment and Cross-Lingual Transfer Emerge in Multilingual Models?}

\author{Tianze Hua$^*$ \\
  Brown University \\
  \texttt{tianze\_hua@brown.edu} \\\And
  Tian Yun$^*$ \\
  Brown University \\
  \texttt{tian\_yun@brown.edu} \\\And
  Ellie Pavlick \\
  Brown University \\
  \texttt{ellie\_pavlick@brown.edu}
}

\begin{document}
\maketitle
\begin{abstract}
\def\thefootnote{*}\footnotetext{Equal contribution.}\def\thefootnote{\arabic{footnote}}
Many pretrained multilingual models exhibit cross-lingual transfer ability, which is often attributed to a learned language-neutral representation during pretraining. However, it remains unclear what factors contribute to the learning of a language-neutral representation, and whether the learned language-neutral representation suffices to facilitate cross-lingual transfer. We propose a synthetic task, Multilingual Othello (mOthello), as a testbed to delve into these two questions. We find that: (1) models trained with naive multilingual pretraining fail to learn a language-neutral representation across all input languages; (2) the introduction of ``anchor tokens'' (i.e., lexical items that are identical across languages) helps cross-lingual representation alignment; and (3) the learning of a language-neutral representation alone is not sufficient to facilitate cross-lingual transfer. Based on our findings, we propose a novel approach -- multilingual pretraining with unified output space -- that both induces the learning of language-neutral representation and facilitates cross-lingual transfer\footnote{All resources will be available at \url{https://github.com/ethahtz/multilingual_othello}}.
\end{abstract}

\section{Introduction}

\begin{figure}[!ht]
    \centering
    \includegraphics[width=0.45\textwidth]{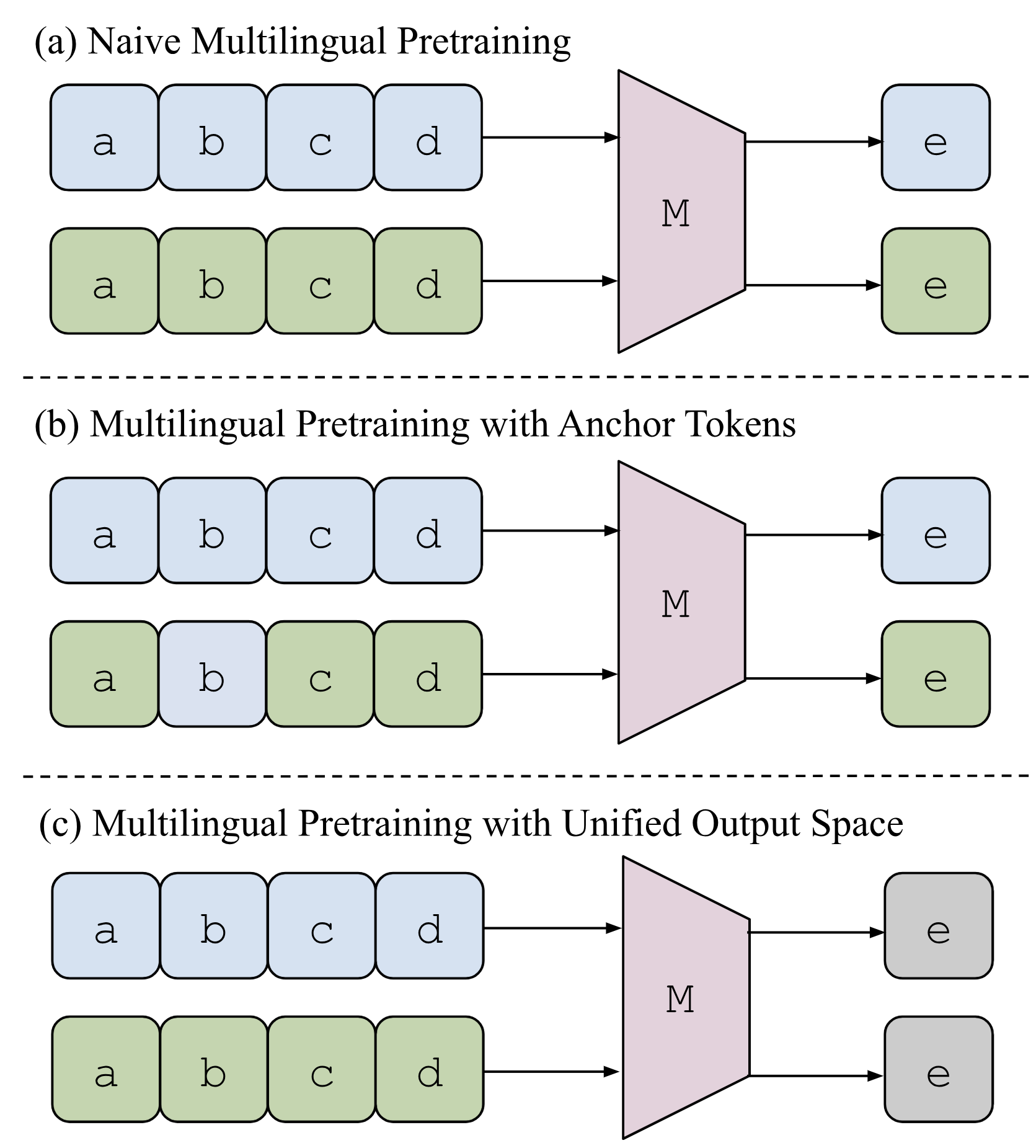}
    \caption{Illustration of three multilingual training approaches. Blue and green blocks represent contexts in 2 different languages, and tokens from the same language have the same color. A multilingual model \texttt{M} consumes \texttt{a,b,c,d} and predicts the corresponding output \texttt{e}. \textbf{Top}: A model is trained on multilingual corpora, with an objective to predict the next tokens specific to each language. \textbf{Middle}: A model is trained on multilingual corpora, where there are tokens shared across language pairs. These tokens are named as anchor tokens. The objective is still to predict the next tokens specific to each language. \textbf{Bottom}: A model is trained on multilingual corpora, with an objective to predict the next tokens in a unified output space.}
    \label{fig: approaches}
\end{figure}

One of the primary desired properties of multilingual models is their cross-lingual transfer ability -- the ability to enhance task performance in a target language when being finetuned exclusively with labeled data from the same task, but in a different source language. Many pretrained multilingual models, such as mBERT \cite{devlin-etal-2019-bert} and XLM-R \cite{conneau-etal-2020-unsupervised}, are found to exhibit this ability across a wide range of tasks, such as natural language inference, named entity recognition, and part-of-speech tagging \cite{pires-etal-2019-multilingual, wu-dredze-2019-beto, KWMR20}. Cross-lingual transfer also serves as a central justification for why we would prefer multilingual models over a collection of monolingual models, despite the fact that the multilingual setting might introduce competition among languages on the model capacity, which has been referred to as ``the curse of multilinguality'' \cite{conneau-etal-2020-unsupervised}.

The cross-lingual transfer ability of pretrained multilingual models is often attributed to a shared, language-neutral space, which is formed during multilingual pretraining \cite{pires-etal-2019-multilingual, libovicky-etal-2020-language, chang-etal-2022-geometry}. However, there is no consensus on what factors lead to such language-neutral representations \cite{wu-dredze-2019-beto, KWMR20, deshpande-etal-2022-bert}. Additionally, to our knowledge, it remains unclear whether a shared, language-neutral space is by itself \textit{sufficient} to facilitate cross-lingual transfer.

In this work, we use a controlled language learning environment to investigate the essential factors for learning language-neutral representations and whether they are sufficient to facilitate the cross-lingual transfer ability of multilingual models. 
To approach these questions, we introduce Multilingual Othello (mOthello), a sequence modeling task based on the Othello board game \cite{li2023emergent}. In mOthello, a model is given a sequence of game moves in a specific ``language'' $\mathcal{L}_k$, and the task is to predict the next legal move in the same ``language'' $\mathcal{L}_k$. This environment is appropriate for our purposes, since it separates the ground truth ``world'' (i.e., the game state) which is assumed to be singular, from the language used to describe it, which can take any number of forms (languages). 

With mOthello, we first train GPT2-based \cite{radford2019language} models (mOthelloGPTs) and analyze the conditions under which language-neutral representations are learned. To quantitatively measure the alignment of representations across languages, we propose cross-lingual alignment probing, which is to recover board states in language $\mathcal{L}_2$ using a probe trained on language $\mathcal{L}_1$. We observe that mOthelloGPTs trained with the naive multilingual pretraining (Figure \ref{fig: approaches}(a)) do not learn language-neutral hidden space across all languages. 

Following, we show that anchor tokens (i.e., shared tokens across languages, Figure \ref{fig: approaches}(b)) facilitates mOthelloGPTs to learn aligned representations across all languages connected via the anchor tokens. However, we observe that these models do not show cross-lingual transfer. This contradicts the common hypothesis that cross-lingual representation alignment suffices for cross-lingual transfer ability of multilingual models. 

Lastly, we further investigate the factors that encourage the emergence of cross-lingual transfer ability. We propose the use of a unified language-neutral output space during multilingual pretraining (Figure \ref{fig: approaches}(c)), which brings both aligned representations across languages and cross-lingual transfer.

\vspace{1cm}
To summarize, our main contributions are: 
\begin{itemize}
    \item We find that models trained with naive multilingual pretraining fail to learn a language-neutral hidden space across all languages.
    \item The introduction of anchor tokens helps cross-lingual representation alignment. 
    \item We observe that the learning of a language-neutral space alone is not sufficient to facilitate cross-lingual transfer. 
    \item We propose an alternative training approach, multilingual pretraining with a unified output space, which both induces the learning of the language-neutral space and facilitates cross-lingual transfer.
\end{itemize}

\section{Related Work}

\subsection{Pretrained Multilingual Models and Cross-lingual Transfer}
Since the success of pretrained English transformer models such as GPT \cite{radford2019language} and BERT \cite{devlin-etal-2019-bert}, there have been interests to replicate this success in the multilingual domain. Multilingual-BERT (mBERT) is trained on a concatenation of monolingual corpora from 104 languages \cite{devlin-etal-2019-bert}, and is found to achieve decent cross-lingual performance and transfer ability \cite{pires-etal-2019-multilingual}. XLM-RoBERTa (XLM-R) \cite{conneau-etal-2020-unsupervised}, with a larger model size and trained on more multilingual data, even achieves on par performance on the GLUE and XNLI task to its monolingual counterparts. 

Cross-lingual transfer refers to the capability of pretrained multilingual models to enhance task performance in a target language when being finetuned exclusively with labeled data from the same task, but in a different source language. There have been extensive work showing the cross-lingual transfer capability of pretrained multilingual models such as mBERT \cite{devlin-etal-2019-bert}, XLM-R \cite{conneau-etal-2020-unsupervised}) and mT5 \cite{xue-etal-2021-mt5}. 

Different hypotheses on the factors associated with a model's ability to transfer across languages have been proposed in previous works, including the amount of shared sub-word tokens across languages \cite{wu-dredze-2019-beto, pires-etal-2019-multilingual, conneau-etal-2020-emerging, KWMR20, deshpande-etal-2022-bert}, typological and structural similarity across languages \cite{pires-etal-2019-multilingual, KWMR20}, comparability of training corpora \cite{dufter-schutze-2020-identifying}, and nature of the task finetuned for cross-lingual transfer (text classification versus text generation) \cite{li-murray-2023-zero}.

In the above works, many have attributed the success of cross-lingual transfer to the language-neutral representations in pretrained multilingual models. However, this hypothesis has not been thoroughly tested, given that the intrinsic characteristics of natural languages impose constraints in the training process of pretrained multilingual models. Our work aims at explicitly testing this hypothesis in a controlled laboratory setting, via the mOthello task, which allows us to have full control over the training data and training approaches.

\subsection{Language-Neutral Representation}

Because of the hypothesized importance of language-neutral representations in cross-lingual transfer, previous works have developed methods to evaluate the extent to which representations of inputs in different languages are language-neutral. These works introduce methods such as measuring the similarity of sentence-level representations of parallel sentences \cite{pires-etal-2019-multilingual, libovicky-etal-2020-language}, conducting statistical analysis of the representational space to separate the language-agnostic and language-specific components \cite{chang-etal-2022-geometry} and investigating token embedding alignment across languages, which is found to be strongly correlated with models' cross-lingual transfer performance \cite{deshpande-etal-2022-bert}. These works all show that in pretrained multilingual models, language-neutral representations are learned. In our work, we want to investigate whether language-neutral representations alone are sufficient for the emergence of  multilingual models' cross-lingual transfer ability.

\subsection{Probing Neural Network Representations}
Probes, typically low-complexity classifiers, have become a standard tool for investigating the information encoded in the hidden representations of language models \citep{alain2016understanding, tenney-et-al2019what, belinkov-glass-2019-analysis}. 
In Othello-GPT \citep{li2023emergent}, states of the game board can be recovered from a GPT \citep{radford2019language} learned to model game moves via trained probes. In this work, we propose cross-lingual alignment probes, which are to reconstruct board states in a target language using a probe trained in a different source language, to quantitatively measure the alignment of representations across languages.

\section{Methods}
\subsection{Othello Game} \label{sec: othello_game_method}
Othello is a strategy board game designed for two players. It is played on an 8x8 grid, totaling 64 tiles. Each player, using either black or white game pieces, takes turns placing a piece on one of the tiles. The game's unique dynamic lies in its limited legal move options at each turn, which involve flipping the opponent's pieces by sandwiching them along a straight line. In the study conducted by \citet{li2023emergent}, the Othello game was transformed into a sequence modeling task. In their adaptation, the model is required to predict the next legal moves based on a sequence of previous moves.

\begin{figure*}[ht]
    \centering
    \begin{subfigure}[t]{0.5\textwidth}
        \centering
        \includegraphics[height=2in]{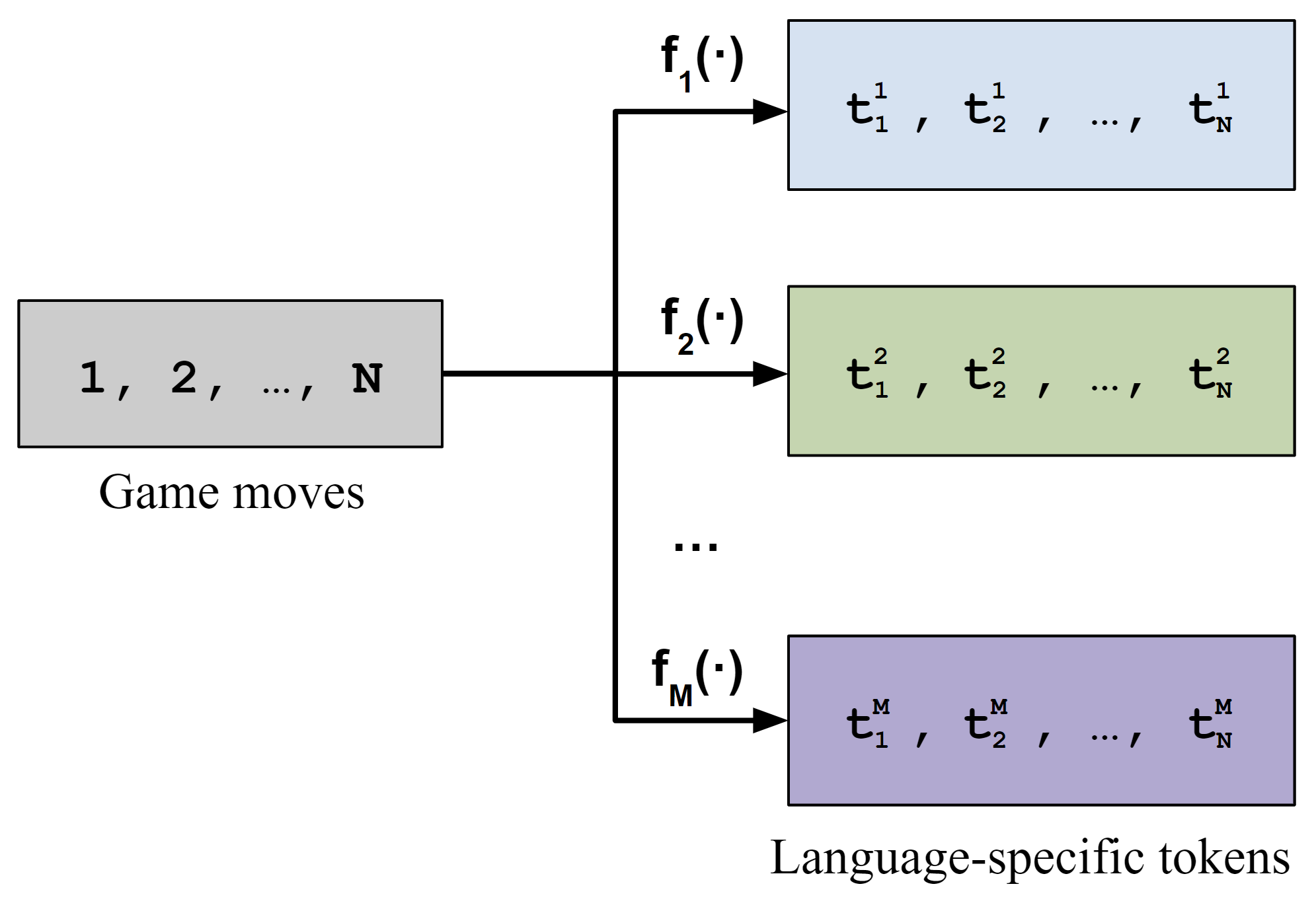}
        \caption{Language-specific Functions}
        \label{subfig: mOthello_funcs}
    \end{subfigure}%
    ~ 
    \begin{subfigure}[t]{0.5\textwidth}
        \centering
        \includegraphics[height=2in]{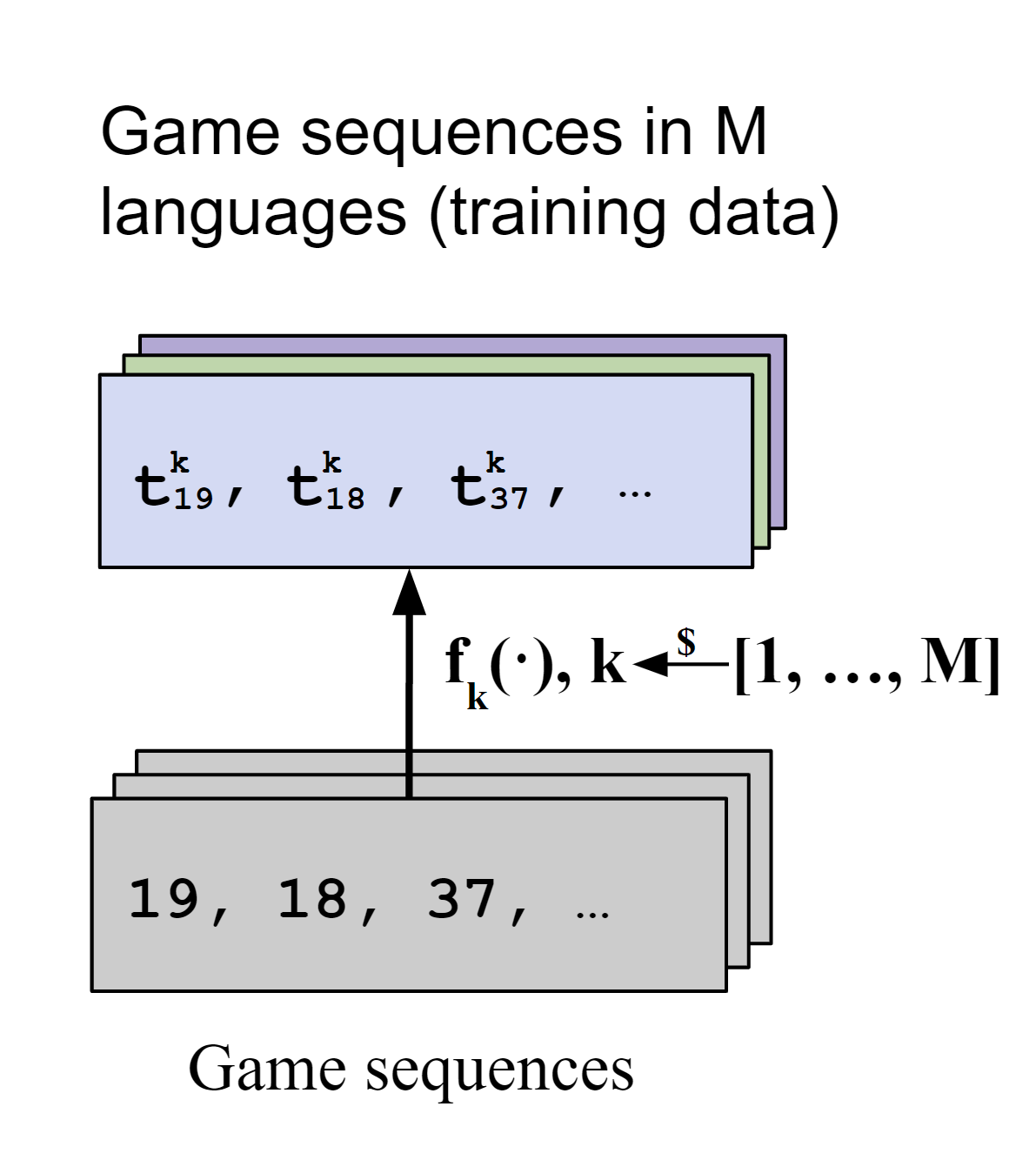}
        \caption{Multilingual Corpus Generation}
        \label{subfig: mOthello_corpus}
    \end{subfigure}
    \caption{An illustration of mOthello. \textbf{Left}: We map game moves to language-specific tokens $t_j^k$ by using a function $f_k$ for language $\mathcal{L}_k$. \textbf{Right}: We create multilingual Othello corpus by mapping Othello game sequences to multilingual Othello language-specific sequences.}
    \label{fig: mOthello}
\end{figure*}

\subsection{Multilingual Othello Game} \label{sec: mOthello_method}
An instance of Multilingual Othello (mOthello) with $M$ languages is defined by\footnote{For simplicity, we assume that there is a one-to-one token-level correspondence across all languages here. We will relax this constraint when we introduce the ``Split'' and ``Compositional'' language variants.}:
\begin{enumerate}
    \item A set of Othello game sequences $S$. For every sequence $s_i = [m_1, m_2, ..., m_{|s_i|}]$, each move $m_j$ (where $1 \leq j \leq |s_i|$) is an integer within the range of $[1,64]$. 
    \item Assume we have $M$ languages. For each language $\mathcal{L}_k$, we define a function $f_k$, which maps each game move to a unique language-specific token. See Figure \ref{subfig: mOthello_funcs} for an illustration of the language-specific functions. A game sequence $s_i$ can be translated into language $\mathcal{L}_k$ by applying $f_k$ on each move in that sequence. The translated sequence in language $\mathcal{L}_k$ can be written as $[f_k(m_j)]_{j=1}^{|s_i|}$. The token space of language $\mathcal{L}_k$ is essentially the range of function $f_k$, which can be noted as [$t_1^k$, ..., $t_{64}^k$]. Note that for all $p, q \in [1, .. M]$, the semantic meaning of tokens $t_j^p$ and $t_j^q$ are the same, since they represent the same underlying move $m_j$. 
\end{enumerate}

Using the functions defined in an instance of mOthello, language-neutral game sequences can be mapped to sequences in different languages. 
The mOthello task is to predict the next legal move in language $\mathcal{L}_k$, given a sequence of previous moves in language $\mathcal{L}_k$. The mOthello task mimics multilingual language modeling, since one not only needs to generate the following natural language tokens, but also generate them in the correct language based on the previous context. 

\subsection{mOthelloGPT \& mOthello Languages} \label{sec: mOthelloGPT_method}
We use the same Transformer-architecture used in \citet{li2023emergent}, which is decoder-only GPT2-style \cite{radford2019language} model. We name this model \textit{mOthelloGPT}. Each mOthelloGPT is trained on $M$ languages, defined by an mOthello instance. 

\begin{figure*}[!ht]
    \centering
    \includegraphics[width=0.8\textwidth]{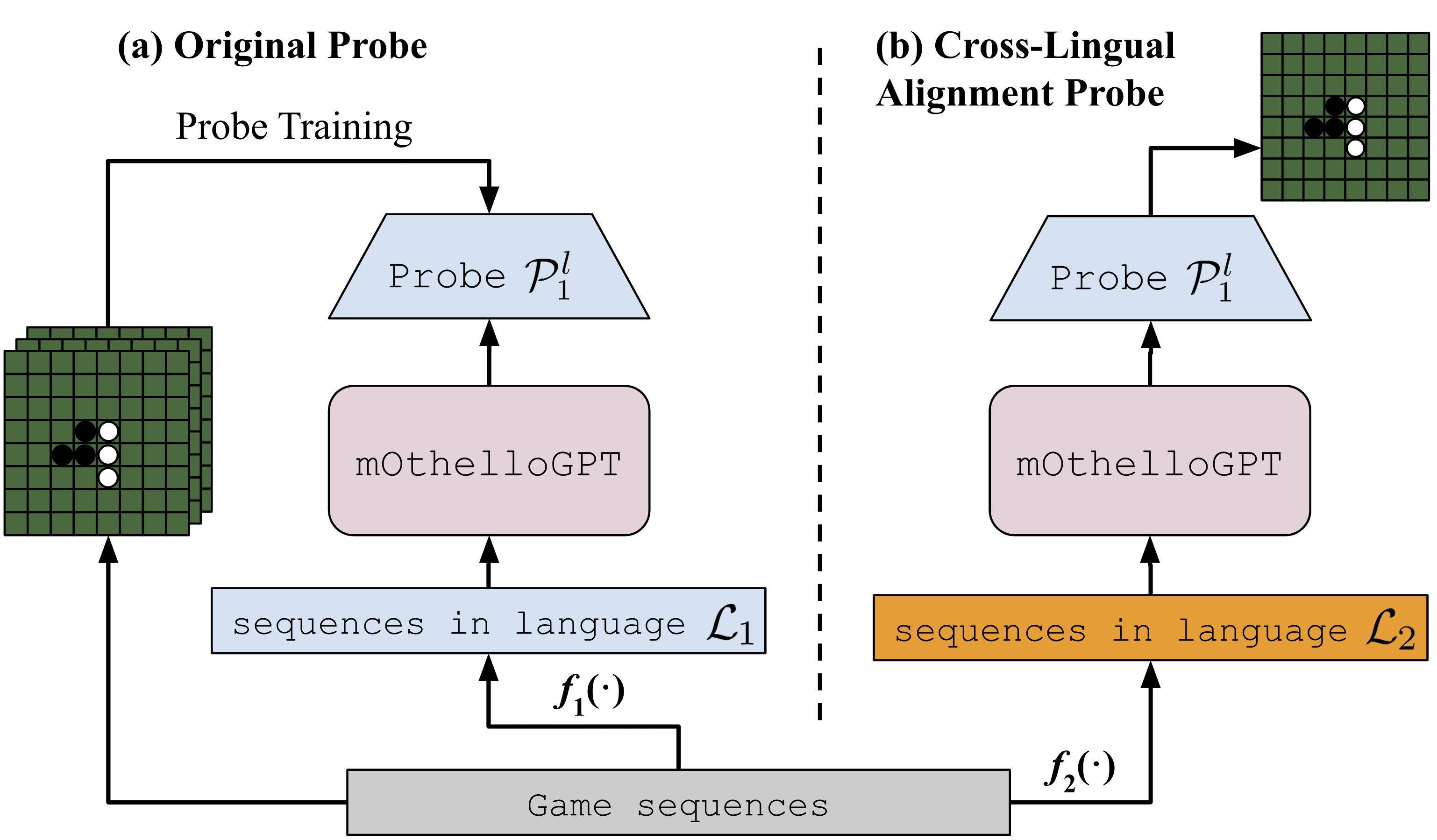}
    \caption{An illustration of the probe training procedure and the cross-lingual alignment probing set-up. Left: we train a probe $\mathcal{P}_1^l$ on the activations at layer $l$ of an mOthelloGPT, using only input sequences in language $\mathcal{L}_1$. The ground-truth labels are obtained by interacting with Othello environment. Right: after probe $\mathcal{P}_1^l$ is trained, we use it to recover the board state given activations at layer $l$ of the same mOthelloGPT model, but using sequences from another language $\mathcal{L}_2$ .}
    \label{fig: transfer_probe}
\end{figure*}

To test the generalizability of our findings beyond the simple mOthello languages, we introduce two variants of mOthello languages to mirror features of natural languages.

\paragraph{The Atomic language} maps each game move to a single (atomic) language-specific token. For example,  moves \texttt{[\textbf{a1}, \textbf{a2}, \textbf{b1}]} are mapped to \texttt{[a1, a2, b1]} in an atomic language.

\paragraph{The Split language} simulates the scenario when a semantic unit is represented by one or more tokens. In the context of mOthello, this means that each game move can be mapped to one or more tokens in a split language. For example, moves \texttt{[\textbf{a1}, \textbf{a2}, \textbf{b1}]} are mapped to \texttt{[a1\textsubscript{1}, a1\textsubscript{2}, a2\textsubscript{1}, b1\textsubscript{1}, b1\textsubscript{2}, b1\textsubscript{3}]} in a split language. The number of tokens each move is split into is sampled randomly from 1 to 3. 

\paragraph{The Compositional language} represents moves by decomposing each of them into its horizontal and vertical location on the board. In this type of language, tokens are reused to represent different moves in a compositional way. For example, moves \texttt{[\textbf{a1}, \textbf{a2}, \textbf{b1}]} are mapped to \texttt{[a, 1, a, 2, b, 1]} in a compositional language. 

Concretely, we investigate whether mOthelloGPTs trained with combinations of Atomic, Split, and Compositional languages can learn language-neutral representations and whether and how cross-lingual transfer ability automatically emerge in these mOthelloGPTs.

\subsection{Cross-lingual Alignment Probes} \label{subsec: transfer_probes}
To investigate to what extent the hidden representations of semantically similar tokens across languages align with one another, we propose \textbf{cross-lingual alignment probes}, which is a probe $\mathcal{P}_{src}^l$ trained to recover the board states with input sequences in language $\mathcal{L}_{src}$ to recover the board states given input sequences in another language $\mathcal{L}_{tgt}$, in a zero-shot fashion. 
If a cross-lingual alignment probe can reconstruct the board states in another language accurately, this reflects that there is a shared latent space for language $\mathcal{L}_{src}$ and $\mathcal{L}_{tgt}$.

To compute cross-lingual alignment probe accuracy from language $\mathcal{L}_1$ to language $\mathcal{L}_2$, we first train the probe on input sequences of $\mathcal{L}_1$. A probe $\mathcal{P}_k^l$ is trained on the activations at layer $l$ of an mOthelloGPT, running on input sequences in language $\mathcal{L}_k$. The input to the probe, $x^l_i$, is a contextualized representation of the $i$-th token in the input sequence at layer $l$ of this mOthelloGPT. We can think of the contextualized token representation as encoding the information of the state of the board after the first $i$ moves in the input sequence. Following this reasoning, the ground-truth labels for training the probe can be computed by running an Othello-simulator on the first $i$ moves. See Figure \ref{fig: transfer_probe} for an illustration of the probe training procedure. 

After probes $\mathcal{P}_1^l$ is trained, we conduct cross-lingual alignment probing by using $\mathcal{P}_1^l$ to recover board states on input sequences in language $\mathcal{L}_2$, as illustrated in Figure \ref{fig: transfer_probe}. A predicted board is then compared with the corresponding ground-truth board. A board consists of 64 tiles, which each can be empty, occupied by a black piece, or occupied by a white piece. 
To calculate the accuracy, we count the number of predicted tiles which match the ground-truth, and divide it by the total number of tiles on the board. For non-atomic languages that represent a move with multiple tokens, we take the contextual representation of the last token of a move for cross-lingual alignment probes.

\section{Experimental Setup}

\subsection{Implementation Details} \label{subsec: m_othello_gpt_datagen}

We use the synthetically generated sequences in \citet{li2023emergent} as the underlying game sequences for the mOthello task.

Figure \ref{subfig: mOthello_corpus} illustrates the corpus generation procedure, which results in the data used to train mOthelloGPTs. First, for each game sequence $s_i$, we randomly select a language $\mathcal{L}_k$. Next, we translate $s_i$ into language $\mathcal{L}_k$ using the corresponding mapping function $f_k$. The resulting translated sequence is then added to the training corpus. 

mOthelloGPT models consist of 8 transformer blocks, each having 8 heads, and a model dimension of 512. mOthelloGPTs are trained on a dataset containing 20 million sequences in $M$ languages with the next-token prediction objective. They are trained for 9 epochs with a batch size of 1024. 

\subsection{Probe Training}
All probes used in this study are two-layer Multi-Layer Perceptrons (MLPs) with a hidden size of 512. Each probe is trained on a set of 800 randomly generated Othello game sequences, which are all translated into a specific language. This results in approximately 48K pairs of activation data and corresponding board states. The probes are trained for 16 epochs, with a batch size of 1024. In this study, probes are trained with activations collected at layer 6 of the mOthelloGPT models, since probes trained with layer 6 activations achieve the highest accuracy compared to probes trained with other layers' activations, providing the highest upper bound for the cross-lingual alignment probe accuracy\footnote{For further details, see Appendix \ref{subsec: layer_to_use}.}. 

\subsection{Cross-lingual Transfer Data}
In cross-lingual transfer experiments, mOthelloGPTs initially undergo pretraining on a dataset of 460K sequences for 40 epochs. This phase includes the use of a 30K-sequence validation dataset for early stopping. These training and validation data for pretraining contain game sequences that all share the same 3 first moves -- this is to ensure that the pretrained models perform sub-optimally on the general game move prediction task, thus leaving space for performance improvement after finetuning. During the finetuning phase, models are finetuned on a smaller dataset containing 102K sequences for 4 epochs. The 102K game sequences in the finetuning data are randomly sampled from the 20 million sequences used in the general model training setup, which include game sequences with arbitrary combinations of the first three moves, thus representing a better distribution. 

\subsection{Cross-lingual Transfer Set-up} \label{subsec: xl_transfer_setup}

We use the following procedure for the cross-lingual transfer experiments: first, we pretrain mOthelloGPTs on a prefix-filtered subset of the Othello corpus\footnote{By prefix-filtered subset, we mean that all the sequences in that subset share the same first few moves. We use prefix-filtered subset as the pretraining corpus because we do not want the pretrained model to generalize too well, hence leaving room for improvement during the finetuning process.}, translated to $M$ languages; then, we finetune the pretrained model with a non-prefix-filtered subset of Othello corpus\footnote{The non-prefix-filtered subset better represents the true distribution of the Othello game sequences.}, but only in one of the languages; finally, we record 5 checkpoints for each epoch of the finetuning process and measure the alignment and performance for each model checkpoint. The performance is measured by calculating the top-1 accuracy of legal move prediction in each language. 
\begin{table}
\centering
\begin{tabular}{@{}lcccc@{}}
\toprule
\textbf{\#Anchor tokens} & \textbf{0} & \textbf{1} & \textbf{2} & \textbf{4} \\ 
\midrule
Atom+Atom & 0.53 & 0.82 & 0.97 & 0.97 \\
Atom+Split & 0.47 & 0.47 & 0.73 & 0.97 \\
Atom+Compositional & 0.51 & 0.59 & 0.67 & 0.97 \\
\bottomrule
\end{tabular}
\caption{Effect of introduced anchor tokens on the alignment of representations across languages in bilingual mOthelloGPTs. Accuracy averaged across 3 different seeds. We find that as the number of introduced anchor tokens increases, the cross-lingual alignment probe accuracy increases, indicating a better alignment of representations across languages.}

\label{table: num_anchor_effect}
\end{table}

\section{Results}

\subsection{Do Cross-Lingual Representations Align under Naive Multilingual Pretraining?} \label{subsec: naive_training_alignment}

We first explore whether hidden representations automatically align across different languages within an mOthelloGPT trained on mOthello sequences. 

The first column of Table \ref{table: num_anchor_effect} shows the pairwise cross-lingual alignment probe accuracy at layer 6 in mOthelloGPTs trained on 3 pairs of languages (i.e., an atomic+atomic language pair, atomic+split language pair and atomic+compositional language pair). We observe that for mOthelloGPTs trained on each of the three pairs of languages, there is a lack of strong alignment in the representations across the languages, implying that naive bilingual pretraining without any inductive biases may not yield representation alignment across languages.

\begin{figure}[ht]
    
    \includegraphics[width=0.49\textwidth]{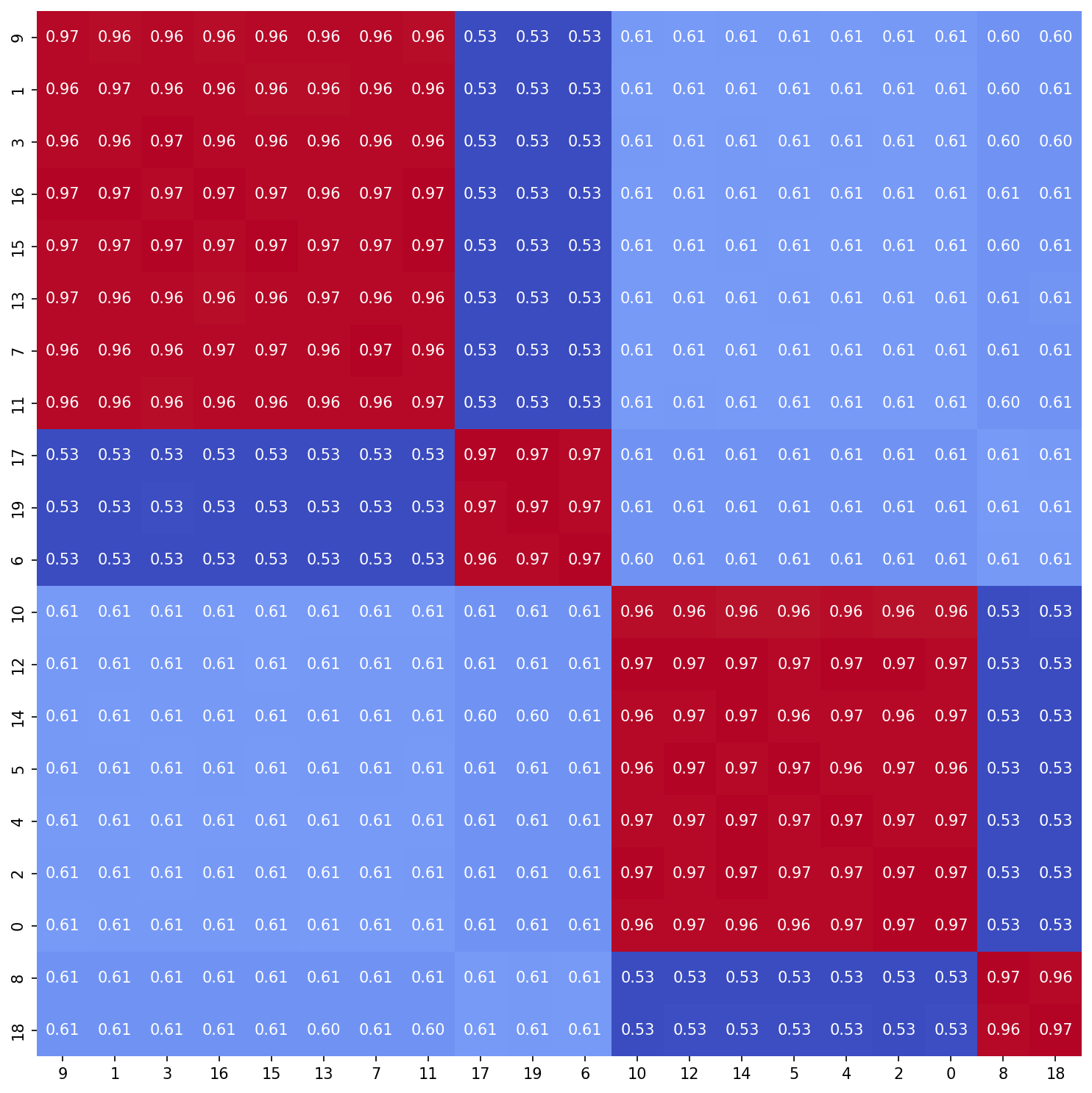}
    \caption{Pairwise cross-lingual alignment probe accuracy for mOthelloGPT trained on 20 atomic languages with naive multilingual pretraining. 
    Each cell $c_{(i,j)}$ reflects the cross-lingual alignment probe accuracy from language $\mathcal{L}_i$ to $\mathcal{L}_j$. For instance, cell $c_{(0,1)}$ indicates the accuracy of board state prediction from input sequences in language $\mathcal{L}_1$ with probe trained on language $\mathcal{L}_0$ to be 0.52. We observe clusters of languages whose representations are aligned with each other, while the alignment of representations across clusters are poor.
    }
    \label{fig: parallel_20}
\end{figure}

Following, we further scale up the bilingual pretraining to multilingual pretraining with 5/10/20/100 languages. Figure \ref{fig: parallel_20} shows the pairwise cross-lingual alignment probe accuracy at layer 6 in mOthelloGPTs trained on 20 atomic languages\footnote{Results for mOthelloGPT trained on 5 and 100 atomic languages can be found in Appendix (Figure \ref{fig: seeding_exp} and Figure \ref{fig: para_100}).}. 
We observe an interesting pattern in models trained with more languages (e.g., 20 and 100): the representations across different languages tend to form clusters. Within these clusters, the accuracy of cross-lingual alignment probes for any pair of languages is high. Conversely, for pairs of languages from different clusters, this accuracy decreases. This pattern demonstrates that some languages may share the same latent space after naive multilingual pretraining, but it is hard to control which set of languages will be aligned together. Despite the formation of language clusters, the misalignment between different clusters reflects that models trained with naive multilingual pretraining are not truly multilingual.

\begin{figure*}[!ht]
    \centering
    \includegraphics[width=1.0\textwidth]{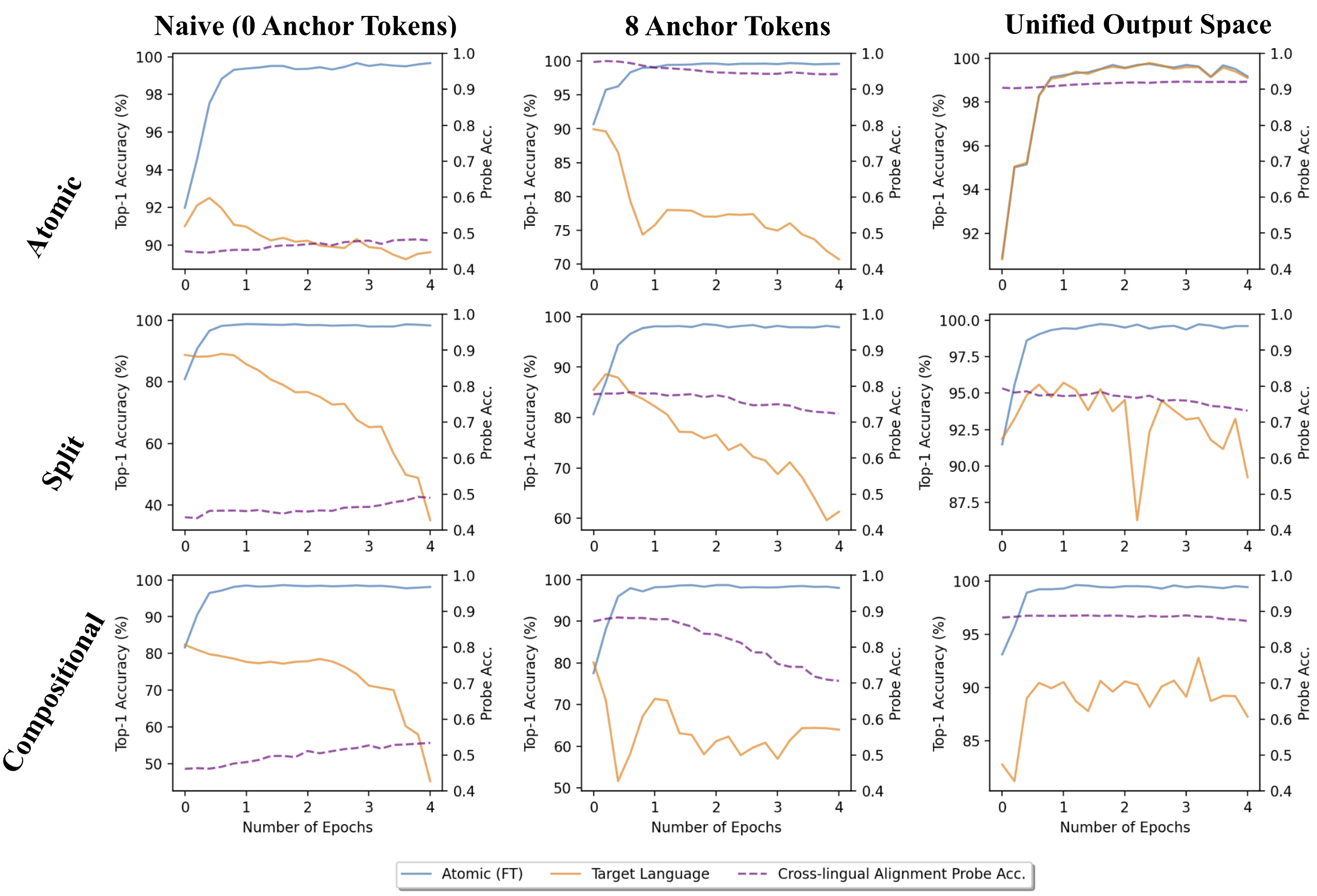}
    \caption{Cross-lingual transfer performance under naive, anchor tokens and unified output space training approaches, of mOthelloGPTs trained on different pairs of languages. Columns (left to right): 1) when 0 anchor tokens are introduced, poor language-neutral representations are learned, which is indicated by the low cross-lingual alignment probe accuracy, 2) when 8 anchor tokens are introduced, rich language-neutral representations are learned in all language pairs, yet cross-lingual transfer performance is poor, indicated by the declining of the target language performance, and 3) when the unified output space approach is taken for training and fine-tuning, we observe that in all language pairs representations are well aligned -- moreover, cross-lingual transfer is also observed, indicated by the improvement of the target language performance.}
    \label{fig: bigfig}
\end{figure*}

\subsection{Multilingual Pretraining with Anchor Tokens Brings Representation Alignment} \label{subsec: anchor_effects}

Multilingual Othello allows us to introduce anchor tokens, which are the shared tokens across languages. 
With anchor tokens, we study their effects on the alignment of cross-lingual representations. 

To approach this, we train mOthelloGPTs on a language pair with different number of anchor tokens and measure the alignment of representations of the language pair. Table \ref{table: num_anchor_effect} shows the averaged cross-lingual transfer probe accuracy based on 3 random seeds for 3 language-pair types.\footnote{For some probing experiments, we observed an unexpected phenomenon and we adjusted our calculation of the probe accuracy. Details can be found in Section \ref{subsec: flipped}.} We observe that as the number of shared anchor tokens across two languages increases, the alignment of representations increases. More specifically, with 4 shared anchor tokens, the representations already reach nearly perfect alignment for all three language-pair types. These observations reflect that anchor tokens significantly helps models to learn aligned representations across languages.

\subsection{Does Cross-Lingual Representation Alignment Facilitate Cross-Lingual Transfer Learning?} \label{subsec: alignment_cross_lingual_transfer_learning}

Next, we study whether aligned cross-lingual representations lead to cross-lingual transfer ability for mOthelloGPTs. 
We conduct cross-lingual transfer experiment on mOthelloGPTs. 
The first and second columns in Figure \ref{fig: bigfig} present cross-lingual transfer results of mOthelloGPTs trained with or without anchor tokens. 
First, we observe that when cross-lingual representations do not align well, mOthelloGPT finetuned on one language does not benefit another language, which means this model does not have cross-lingual transfer ability.
Surprisingly, we find that even when the cross-lingual representation alignment is high for an mOthelloGPT, cross-lingual transfer still does not occur.
This finding goes against the common belief that cross-lingual representation alignment is a sufficient condition for the emergence of cross-lingual transfer ability in multilingual models.

\subsection{Multilingual Pretraining with Unified Output Space Brings Representation Alignment and Cross-Lingual Transfer Ability} \label{subsec: unified}

So far, we have seen that the alignment of representations is not sufficient to guarantee cross-lingual transfer learning across languages. 

Inspired by methods proposed to improve cross-lingual transfer via intermediate-task training \cite{phang-etal-2020-english} and language-independent entity prediction task training \cite{calixto-etal-2021-wikipedia}, we introduce multilingual pretraining with unified output space to facilitate cross-lingual representation alignment and cross-lingual transfer ability.
Specifically, we train an mOthelloGPT which consumes sequence in two source languages, $\mathcal{L}_\text{src1}$ and $\mathcal{L}_\text{src2}$, and predicts sequences in a unified output space, noted as $\mathcal{L}_\text{tgt}$. 
We then measure representation alignment in the two source languages, as well as model's cross-lingual transfer ability.

The third column in Figure \ref{fig: bigfig} shows the results of representation alignment and cross-lingual transfer learning under the multilingual pretraining with unified output space.
We observe that pretraining with unified output space brings mOthelloGPTs not only cross-lingual alignment, but also cross-lingual transfer ability. 
Specifically, for mOthelloGPT pretrained with Atomic language pairs, the cross-lingual alignment probe accuracy remains at around 90\%, indicating that $\mathcal{L}_\text{src1}$ and $\mathcal{L}_\text{src2}$ are well aligned. Moreover, we observe that despite not encountering any sequences from language $\mathcal{L}_\text{src2}$ during finetuning, this mOthelloGPT still manages to enhance its performance in predicting next legal moves in language $\mathcal{L}_\text{src2}$ to the same extent as in language $\mathcal{L}_\text{src1}$. This indicates that this mOthelloGPT achieves cross-lingual transfer under the unified output space approach. 
We notice that the cross-lingual transfer ability of mOthelloGPTs trained with Split or Compositional language pairs is slightly weaker, but the pattern that finetuning on $\mathcal{L}_\text{src1}$ benefits next move prediction in $\mathcal{L}_\text{src2}$ still holds, especially at early finetuning phase.

The improvement in performance of $\mathcal{L}_\text{src2}$ across three language pairs of structurally different languages implies that multilingual pretraining with unified output space is an effective approach for inducing cross-lingual alignment and cross-lingual transfer ability and is robust to structural differences across languages.

\begin{figure*}[!ht]
    \centering
    \includegraphics[width=1.0\textwidth]{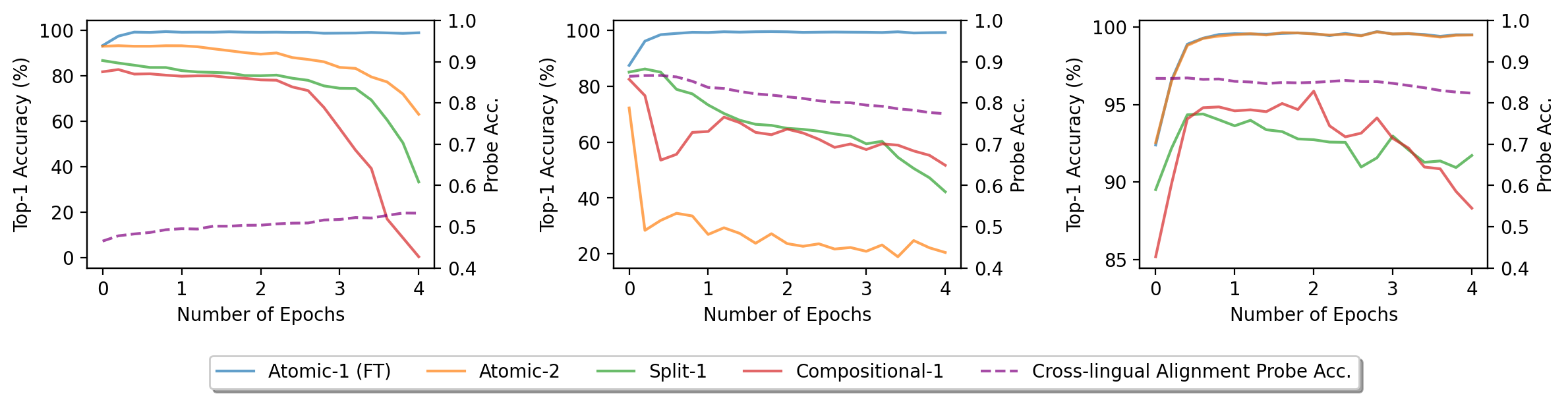}
    \caption{Cross-lingual transfer performance under naive, anchor tokens and unified output space training approaches, of mOthelloGPTs trained on 4 languages consisted of multiple types. Cross-lingual alignment probe accuracy is computed as the average of the accuracy between the finetuning language and each of the target languages. Under the naive training (left figure), the average cross-lingual alignment probe accuracy is low and the improvement of the finetuning language does not transfer to the target languages; under the introduction of anchor tokens, although the average representation alignment is better, still no cross-lingual transfer is found; under the unified output space approach, both well aligned representation and cross-lingual transfer are observed.}
    \label{fig: mixed_ft}
\end{figure*}

\subsection{Multilingual Pretraining with More than Two Languages}
In previous sections, we explored how different training approaches affect alignment of representations and cross-lingual transfer in bilingual mOthelloGPT models. Here, we explore whether our findings hold for multilingual models that are trained with more than two languages. 
Figure \ref{fig: mixed_ft} shows the cross-lingual representation alignment and cross-lingual transfer performance of mOthelloGPTs trained with 4 languages consisting of different language types. 
We find that the results are consistent with our findings on bilingual mOthelloGPTs: (1) While anchor tokens improve representation alignment across languages, it does not help the model to achieve its cross-lingual transfer ability; (2) With the introduction of the unified output token space during multilingual pretraining, both cross-lingual representation alignment and cross-lingual transfer are achieved. This result suggests that the unified output space approach also generalizes to scenarios when a multilingual model is trained on more than two languages.
\section{Discussions}

In Section \ref{subsec: unified}, we introduced the unified output space approach to induce both representation alignment and cross-lingual transfer in mOthelloGPTs. However, it is important to note that modeling mOthello is considerably simpler than modeling natural languages. While it is simple to identify a next token in the unified output space in mOthello, it is comparatively challenging to identify a language-neutral next token for each language-specific context in natural languages. Nevertheless, our results could inform future strategies in designing training objectives for multilingual models. Traditional training of such models primarily focuses on the task of predicting the next language-specific token. Our results suggest that incorporating an auxiliary language-neutral prediction task into the training process could potentially enhance the cross-lingual transfer abilities of multilingual models. Some such approaches have been explored in \cite{phang-etal-2020-english, calixto-etal-2021-wikipedia}.

\section{Conclusion}
In this paper, we propose the Multilingual Othello (mOthello) sequence modeling task as a testbed to investigate the factors which help align representations across languages, and to study the relationship between representation alignment and cross-lingual transfer in multilingual models. We introduce a new metric, the cross-lingual alignment probe accuracy, on measuring the alignment of representations across languages. We train mOthelloGPTs on the mOthello task, and conducted analyses on the representation alignment. We found that models trained with a naive approach fail to learn a language-neutral hidden space across all input languages, but the introduction of anchor tokens helps the alignment of representations. Then, we conduct finetuning experiments on mOthelloGPTs pretrained on a prefix-limited training corpus. To our surprise, we found that the learning of a language-neutral space alone is not sufficient to facilitate cross-lingual transfer. Upon further investigation, we propose an alternative training approach -- the unified output space approach -- that both induces the learning of the language-neutral space and facilitates cross-lingual transfer.

\section*{Limitations}
Our study used the toy task mOthello and its synthetic variants to investigate the alignment of representations across languages and the cross-lingual transfer ability in multilingual models. However, it is important to note that in real-world scenarios, the vocabulary size of each language is substantially larger than the token space in mOthello, which contains only less than 180 tokens per language. Additionally, our experiments were conducted on models with a decoder-only transformer architecture. This focus leaves out a significant portion of state-of-the-art multilingual models, many of which employ encoder-decoder and encoder-only architectures. These factors should be considered when interpreting the applicability of our findings to broader, more complex linguistic contexts. 

\bibliography{custom}

\begin{thebibliography}{19}
\expandafter\ifx\csname natexlab\endcsname\relax\def\natexlab#1{#1}\fi

\bibitem[{Alain and Bengio(2017)}]{alain2016understanding}
Guillaume Alain and Yoshua Bengio. 2017.
\newblock \href {https://openreview.net/forum?id=HJ4-rAVtl} {Understanding intermediate layers using linear classifier probes}.
\newblock In \emph{The Fifth International Conference on Learning Representations, Workshop Track Proceedings}.

\bibitem[{Belinkov and Glass(2019)}]{belinkov-glass-2019-analysis}
Yonatan Belinkov and James Glass. 2019.
\newblock \href {https://doi.org/10.1162/tacl_a_00254} {Analysis methods in neural language processing: A survey}.
\newblock \emph{Transactions of the Association for Computational Linguistics}, 7:49--72.

\bibitem[{Calixto et~al.(2021)Calixto, Raganato, and Pasini}]{calixto-etal-2021-wikipedia}
Iacer Calixto, Alessandro Raganato, and Tommaso Pasini. 2021.
\newblock \href {https://doi.org/10.18653/v1/2021.naacl-main.286} {{W}ikipedia entities as rendezvous across languages: Grounding multilingual language models by predicting {W}ikipedia hyperlinks}.
\newblock In \emph{Proceedings of the 2021 Conference of the North American Chapter of the Association for Computational Linguistics: Human Language Technologies}, pages 3651--3661, Online. Association for Computational Linguistics.

\bibitem[{Chang et~al.(2022)Chang, Tu, and Bergen}]{chang-etal-2022-geometry}
Tyler Chang, Zhuowen Tu, and Benjamin Bergen. 2022.
\newblock \href {https://doi.org/10.18653/v1/2022.emnlp-main.9} {The geometry of multilingual language model representations}.
\newblock In \emph{Proceedings of the 2022 Conference on Empirical Methods in Natural Language Processing}, pages 119--136, Abu Dhabi, United Arab Emirates. Association for Computational Linguistics.

\bibitem[{Conneau et~al.(2020{\natexlab{a}})Conneau, Khandelwal, Goyal, Chaudhary, Wenzek, Guzm{\'a}n, Grave, Ott, Zettlemoyer, and Stoyanov}]{conneau-etal-2020-unsupervised}
Alexis Conneau, Kartikay Khandelwal, Naman Goyal, Vishrav Chaudhary, Guillaume Wenzek, Francisco Guzm{\'a}n, Edouard Grave, Myle Ott, Luke Zettlemoyer, and Veselin Stoyanov. 2020{\natexlab{a}}.
\newblock \href {https://doi.org/10.18653/v1/2020.acl-main.747} {Unsupervised cross-lingual representation learning at scale}.
\newblock In \emph{Proceedings of the 58th Annual Meeting of the Association for Computational Linguistics}, pages 8440--8451, Online. Association for Computational Linguistics.

\bibitem[{Conneau et~al.(2020{\natexlab{b}})Conneau, Wu, Li, Zettlemoyer, and Stoyanov}]{conneau-etal-2020-emerging}
Alexis Conneau, Shijie Wu, Haoran Li, Luke Zettlemoyer, and Veselin Stoyanov. 2020{\natexlab{b}}.
\newblock \href {https://doi.org/10.18653/v1/2020.acl-main.536} {Emerging cross-lingual structure in pretrained language models}.
\newblock In \emph{Proceedings of the 58th Annual Meeting of the Association for Computational Linguistics}, pages 6022--6034, Online. Association for Computational Linguistics.

\bibitem[{Deshpande et~al.(2022)Deshpande, Talukdar, and Narasimhan}]{deshpande-etal-2022-bert}
Ameet Deshpande, Partha Talukdar, and Karthik Narasimhan. 2022.
\newblock \href {https://doi.org/10.18653/v1/2022.naacl-main.264} {When is {BERT} multilingual? isolating crucial ingredients for cross-lingual transfer}.
\newblock In \emph{Proceedings of the 2022 Conference of the North American Chapter of the Association for Computational Linguistics: Human Language Technologies}, pages 3610--3623, Seattle, United States. Association for Computational Linguistics.

\bibitem[{Devlin et~al.(2019)Devlin, Chang, Lee, and Toutanova}]{devlin-etal-2019-bert}
Jacob Devlin, Ming-Wei Chang, Kenton Lee, and Kristina Toutanova. 2019.
\newblock \href {https://doi.org/10.18653/v1/N19-1423} {{BERT}: Pre-training of deep bidirectional transformers for language understanding}.
\newblock In \emph{Proceedings of the 2019 Conference of the North {A}merican Chapter of the Association for Computational Linguistics: Human Language Technologies, Volume 1 (Long and Short Papers)}, pages 4171--4186, Minneapolis, Minnesota. Association for Computational Linguistics.

\bibitem[{Dufter and Sch{\"u}tze(2020)}]{dufter-schutze-2020-identifying}
Philipp Dufter and Hinrich Sch{\"u}tze. 2020.
\newblock \href {https://doi.org/10.18653/v1/2020.emnlp-main.358} {Identifying elements essential for {BERT}{'}s multilinguality}.
\newblock In \emph{Proceedings of the 2020 Conference on Empirical Methods in Natural Language Processing (EMNLP)}, pages 4423--4437, Online. Association for Computational Linguistics.

\bibitem[{K et~al.(2020)K, Wang, Mayhew, and Roth}]{KWMR20}
Karthikeyan K, Zihan Wang, Stephen Mayhew, and Dan Roth. 2020.
\newblock \href {https://openreview.net/forum?id=HJeT3yrtDr} {Cross-lingual ability of multilingual bert: An empirical study}.
\newblock In \emph{The Eighth International Conference on Learning Representations}.

\bibitem[{Li et~al.(2023)Li, Hopkins, Bau, Vi{\'e}gas, Pfister, and Wattenberg}]{li2023emergent}
Kenneth Li, Aspen~K Hopkins, David Bau, Fernanda Vi{\'e}gas, Hanspeter Pfister, and Martin Wattenberg. 2023.
\newblock \href {https://openreview.net/forum?id=DeG07_TcZvT} {Emergent world representations: Exploring a sequence model trained on a synthetic task}.
\newblock In \emph{The Eleventh International Conference on Learning Representations}.

\bibitem[{Li and Murray(2023)}]{li-murray-2023-zero}
Tianjian Li and Kenton Murray. 2023.
\newblock \href {https://doi.org/10.18653/v1/2023.findings-acl.789} {Why does zero-shot cross-lingual generation fail? an explanation and a solution}.
\newblock In \emph{Findings of the Association for Computational Linguistics: ACL 2023}, pages 12461--12476, Toronto, Canada. Association for Computational Linguistics.

\bibitem[{Libovick{\'y} et~al.(2020)Libovick{\'y}, Rosa, and Fraser}]{libovicky-etal-2020-language}
Jind{\v{r}}ich Libovick{\'y}, Rudolf Rosa, and Alexander Fraser. 2020.
\newblock \href {https://doi.org/10.18653/v1/2020.findings-emnlp.150} {On the language neutrality of pre-trained multilingual representations}.
\newblock In \emph{Findings of the Association for Computational Linguistics: EMNLP 2020}, pages 1663--1674, Online. Association for Computational Linguistics.

\bibitem[{Phang et~al.(2020)Phang, Calixto, Htut, Pruksachatkun, Liu, Vania, Kann, and Bowman}]{phang-etal-2020-english}
Jason Phang, Iacer Calixto, Phu~Mon Htut, Yada Pruksachatkun, Haokun Liu, Clara Vania, Katharina Kann, and Samuel~R. Bowman. 2020.
\newblock \href {https://aclanthology.org/2020.aacl-main.56} {{E}nglish intermediate-task training improves zero-shot cross-lingual transfer too}.
\newblock In \emph{Proceedings of the 1st Conference of the Asia-Pacific Chapter of the Association for Computational Linguistics and the 10th International Joint Conference on Natural Language Processing}, pages 557--575, Suzhou, China. Association for Computational Linguistics.

\bibitem[{Pires et~al.(2019)Pires, Schlinger, and Garrette}]{pires-etal-2019-multilingual}
Telmo Pires, Eva Schlinger, and Dan Garrette. 2019.
\newblock \href {https://doi.org/10.18653/v1/P19-1493} {How multilingual is multilingual {BERT}?}
\newblock In \emph{Proceedings of the 57th Annual Meeting of the Association for Computational Linguistics}, pages 4996--5001, Florence, Italy. Association for Computational Linguistics.

\bibitem[{Radford et~al.(2019)Radford, Wu, Child, Luan, Amodei, Sutskever et~al.}]{radford2019language}
Alec Radford, Jeffrey Wu, Rewon Child, David Luan, Dario Amodei, Ilya Sutskever, et~al. 2019.
\newblock Language models are unsupervised multitask learners.
\newblock \emph{OpenAI blog}, 1(8):9.

\bibitem[{Tenney et~al.(2019)Tenney, Xia, Chen, Wang, Poliak, McCoy, Kim, Durme, Bowman, Das, and Pavlick}]{tenney-et-al2019what}
Ian Tenney, Patrick Xia, Berlin Chen, Alex Wang, Adam Poliak, R.~Thomas McCoy, Najoung Kim, Benjamin~Van Durme, Samuel~R. Bowman, Dipanjan Das, and Ellie Pavlick. 2019.
\newblock \href {https://openreview.net/forum?id=SJzSgnRcKX} {What do you learn from context? probing for sentence structure in contextualized word representations}.
\newblock In \emph{The Sixth International Conference on Learning Representations}.

\bibitem[{Wu and Dredze(2019)}]{wu-dredze-2019-beto}
Shijie Wu and Mark Dredze. 2019.
\newblock \href {https://doi.org/10.18653/v1/D19-1077} {Beto, bentz, becas: The surprising cross-lingual effectiveness of {BERT}}.
\newblock In \emph{Proceedings of the 2019 Conference on Empirical Methods in Natural Language Processing and the 9th International Joint Conference on Natural Language Processing (EMNLP-IJCNLP)}, pages 833--844, Hong Kong, China. Association for Computational Linguistics.

\bibitem[{Xue et~al.(2021)Xue, Constant, Roberts, Kale, Al-Rfou, Siddhant, Barua, and Raffel}]{xue-etal-2021-mt5}
Linting Xue, Noah Constant, Adam Roberts, Mihir Kale, Rami Al-Rfou, Aditya Siddhant, Aditya Barua, and Colin Raffel. 2021.
\newblock \href {https://doi.org/10.18653/v1/2021.naacl-main.41} {m{T}5: A massively multilingual pre-trained text-to-text transformer}.
\newblock In \emph{Proceedings of the 2021 Conference of the North American Chapter of the Association for Computational Linguistics: Human Language Technologies}, pages 483--498, Online. Association for Computational Linguistics.

\end{thebibliography}

\newpage
\appendix
\section{Appendix}

\begin{figure*}[ht]
    \centering
    \includegraphics[width=\textwidth]{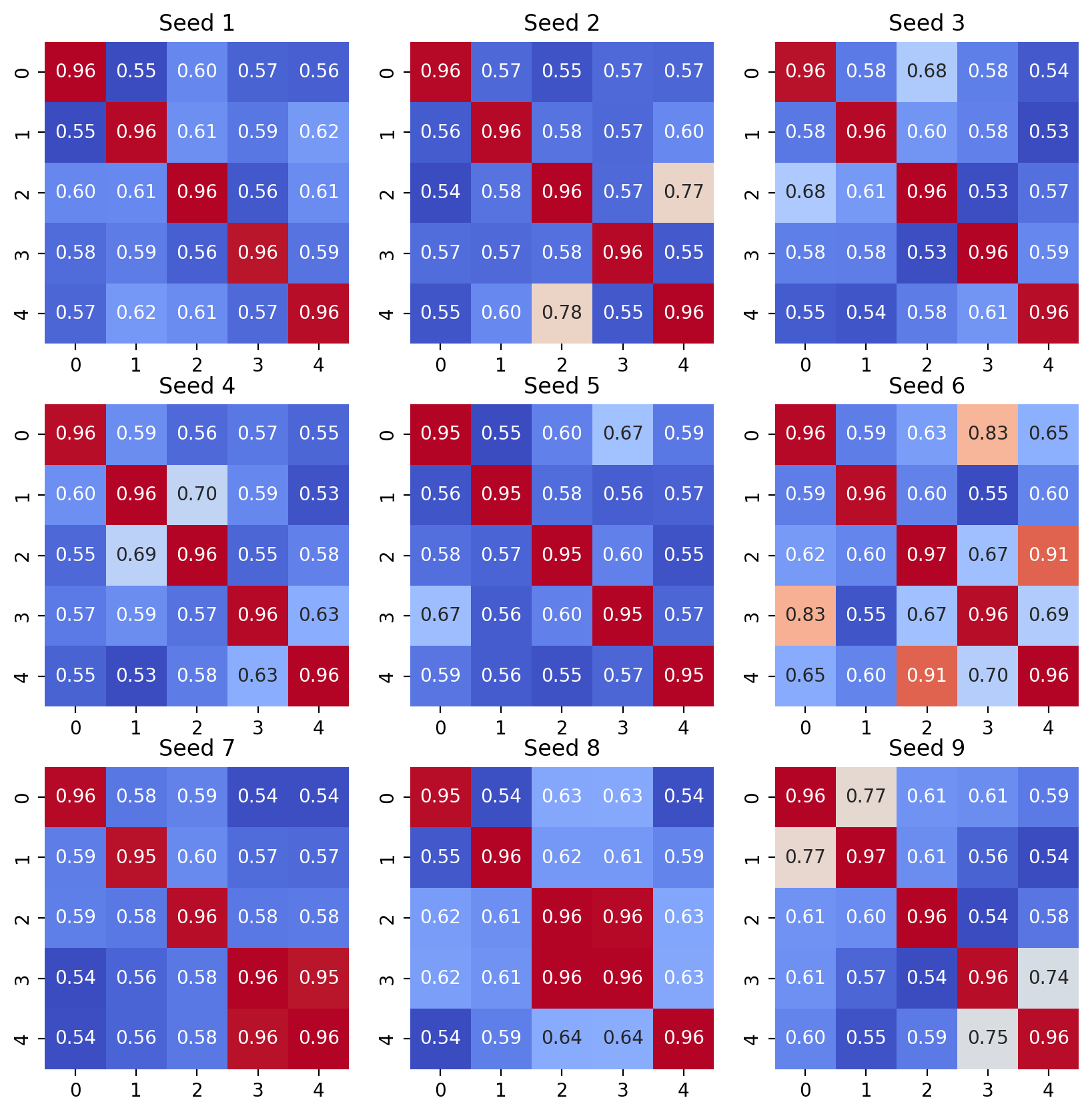}
    \caption{Pairwise cross-lingual alignment probe accuracy for mOthelloGPTs initialized with 9 different seeds, each trained on sequences from 5 atomic languages.}
    \label{fig: seeding_exp}
\end{figure*}

\begin{figure*}[ht]
    \centering
    \includegraphics[width=\textwidth]{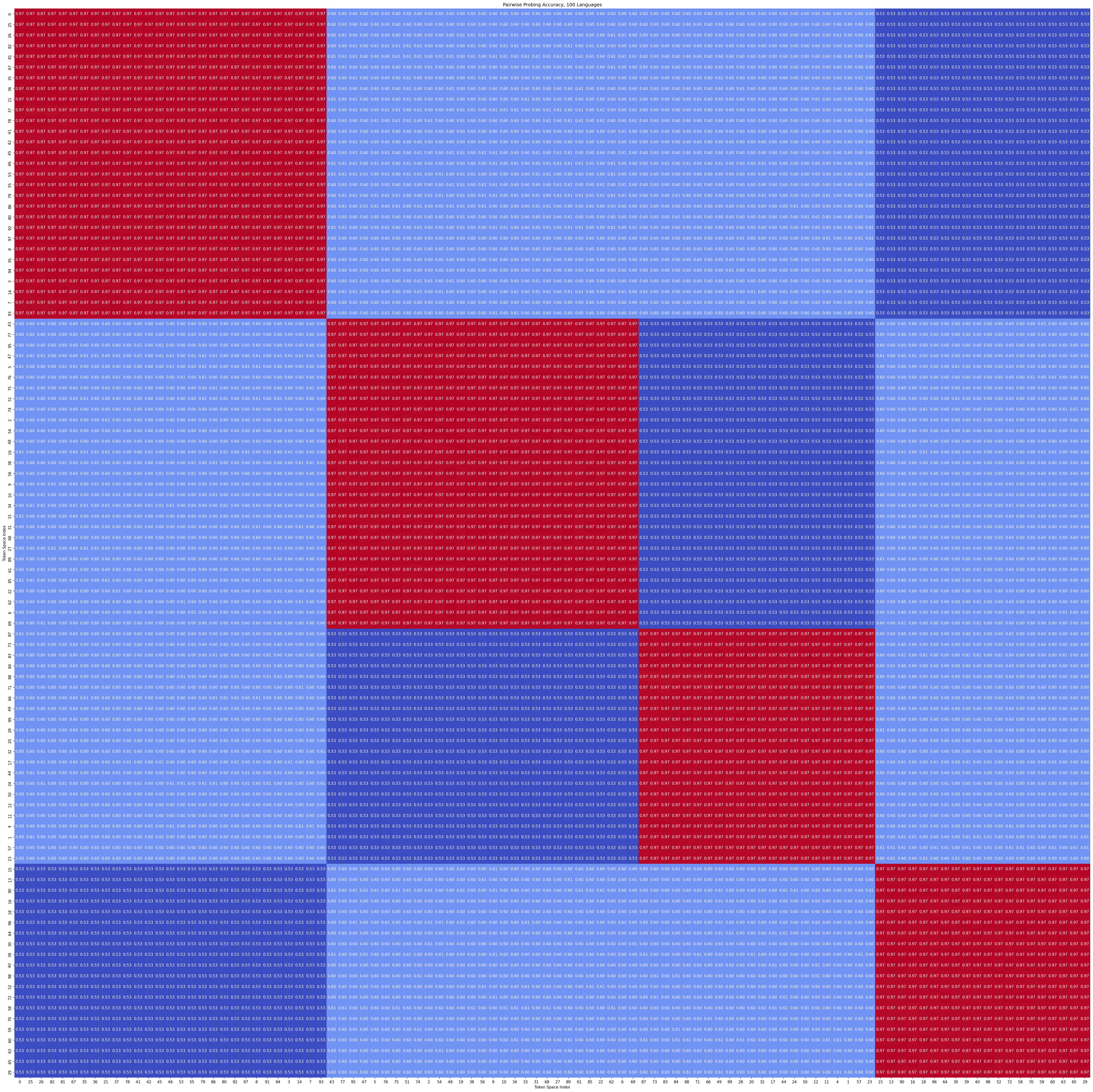}
    \caption{Pairwise cross-lingual alignment probe accuracy for an mOthelloGPT trained on sequences from 100 atomic languages.}
    \label{fig: para_100}
\end{figure*}

\subsection{Layer Choice for Cross-lingual Alignment Probes} \label{subsec: layer_to_use} 
Not all layers are suitable for training probes and computing cross-lingual alignment probe accuracy. In the Othello work, the authors found that the probes are particularly good at extracting board states when trained on activations from layer 5 to layer 7. To select a layer for training probes, we have computed the original and cross-lingual alignment probe accuracy across all layers in Table \ref{table: choice_layer6} for a bilingual mOthelloGPT trained naively. We chose layer 6 for our study because it exhibits the highest original probe accuracy (i.e. the probe is trained and tested on sequences from a same language). This original probe accuracy serves as an approximate upper-bound for the cross-lingual alignment probe accuracy, i.e. the cross-lingual alignment probe do not outperform the original probe in its prediction accuracy. Our aim is to ensure this upper-bound accuracy is as high as possible such that: if we observe low cross-lingual alignment probe accuracy, it suggests that the issue is not due to an inherent inability for any probe to accurately predict the board state, but rather that the representations between languages are unaligned, thereby showcasing the low performance of the cross-lingual alignment probe. This distinction is crucial for correctly interpreting the implications of low cross-lingual alignment probe accuracy.

\begin{table*}
\centering
\begin{tabular}{@{}lccccccccc@{}}
\toprule
\textbf{Layer index} & \textbf{0} & \textbf{1} & \textbf{2} & \textbf{3} & \textbf{4} & \textbf{5} & \textbf{6} & \textbf{7} & \textbf{8} \\ 
\midrule
Original Probe acc. & 0.539 &  0.837 &  0.892 &  0.932 &  0.953 &  0.963 &  0.965 &  0.953 &  0.769 \\
Cross Probe acc. & 0.486 &  0.570 &  0.578 &  0.570 &  0.552 &  0.537 &  0.527 &  0.516 &  0.468 \\
\bottomrule
\end{tabular}
\caption{Original probe accuracy and cross-lingual alignment probe accuracy computed using probes trained across all layers from a bilingual mOthelloGPT.}
\label{table: choice_layer6}
\end{table*}

\subsection{Validating Probes through Intervention Experiments}
To see whether the probes trained in this study are extracting causally significant board states with regard to the model's legal move predictions, we run the same intervention analysis as in \citet{li2023emergent}. From Table \ref{table: intervention_res}, we can see that probes are capable of intervening with the model's internal representation and alter its legal move predictions based on the edited game board state. The cross-lingual alignment probe, when used to intervene the board states given inputs from a different but aligned language, can perform almost as good as the probe that has been originally trained on that language.

\begin{table}
\centering
\begin{tabular}{@{}lcc@{}}
\toprule
~ & \textbf{Non-Aligned} & \textbf{Aligned}  \\ 
\midrule
Original Probe & 0.28 &  0.19 \\
Cross Probe & 2.75 &  0.20 \\
Null \cite{li2023emergent} & \multicolumn{2}{c}{2.68}\\
\bottomrule
\end{tabular}
\caption{Average error using original probe and cross-lingual alignment probe on mOthelloGPTs that learned non-aligned or aligned language representations. In the non-aligned group, using the original probe leads to a much lower intervention error, while intervening with the cross-lingual alignment probe leads to an error rate no better than null-intervention baseline. In the aligned group, we found that the intervention error using the original probe and the cross-lingual alignment probe are similar, and both significantly outperforms the null baseline.}
\label{table: intervention_res}
\end{table}

\subsection{Effect of Random Initialization on Representation Alignment}
We also delve into the impact of random initialization on naive multilingual pretraining. Based on Figure \ref{fig: seeding_exp}, we observe that language clusters sometimes may appear, but there is no perfect alignment across all languages, which serves as another evidence that naive multilingual pretraining does not yield alignment of language representation. 

\subsection{Indirect Effect of Anchor Tokens}
To explore the indirect effect of anchor tokens, we focus on mOthelloGPT models trained on three atomic languages for which two out of three pairs of languages share some anchor tokens. For instance, consider the set of languages $\mathcal{L}_1$, $\mathcal{L}_2$, and $\mathcal{L}_3$ as an example. We make $\mathcal{L}_1$ and $\mathcal{L}_2$ share some anchor tokens, and $\mathcal{L}_2$ and $\mathcal{L}_3$ share some other anchor tokens. We select the anchor tokens in a way such that languages $\mathcal{L}_1$ and $\mathcal{L}_3$ have completely disjoint token spaces so that we can explore the indirect effect of anchor tokens. Results are shown in Table \ref{table: indirect_anchor_exp}, showing a significant indirect effect of anchor tokens on language pairs that do not share any anchor tokens directly. 

\subsection{Color-Flipped State Predictions of the Cross-lingual Alignment Probes}
\label{subsec: flipped}
During the process of measuring cross-lingual alignment probe accuracy for Table \ref{table: num_anchor_effect}, we found that some cross-lingual alignment probes, when being use to predict board states in a different split or compositional language, predict the near-perfect color-flipped state of the game board (i.e. the black pieces are predicted to be white pieces, and the white pieces black pieces). On a representation level, it is reasonable to argue that mOthelloGPTs still learn a shared representation of the game board across languages even if the probe predicts a color-flipped state of the board, since a color-flipped state of the board encodes exactly the same information as its counterpart, and the probe's prediction of black and white pieces is subject to the arbitrarily chosen labels during the training of the probe. Therefore, to better capture a more flexible notion of board-state representation, we take the maximum of the plain probe prediction accuracy and the color-flipped probe prediction accuracy as the final cross-lingual alignment probe accuracy. This change does not affect cross-lingual alignment probe accuracy for unaligned representations since even if the predicted colors of all pieces are flipped, the resulting accuracy still will be no better than the plain prediction accuracy.

\begin{table}
\centering
\begin{tabular}{@{}lccc@{}}
\toprule
 & \multicolumn{3}{c}{Language Pairs} \\
\cmidrule(r){2-4}
 & (0,1) & (0,2) & (1,2) \\
\midrule
\multicolumn{4}{l}{\textbf{2 anchor tokens per (0,1) and (0,2)}} \\
Cross Probe Acc. & 0.91 & 0.97 & 0.90 \\
\midrule
\multicolumn{4}{l}{\textbf{4 anchor tokens per (0,1) and (0,2)}} \\
Cross Probe Acc. & 0.97 & 0.97 & 0.97 \\
\bottomrule
\end{tabular}
\caption{Indirect effects of anchor tokens. For each experiment, the third column shows the extent to which the representations of language $\mathcal{L}_1$ and language $\mathcal{L}_2$ align with each other. Whenever the representations of language pairs ($\mathcal{L}_0$,$\mathcal{L}_1$) and ($\mathcal{L}_0$,$\mathcal{L}_2$) are aligned, the representations between languages $\mathcal{L}_1$ and $\mathcal{L}_2$ are also aligned. The second experiment illustrates an example where language pair ($\mathcal{L}_0$,$\mathcal{L}_1$) is less aligned, which led to language pair ($\mathcal{L}_1$, $\mathcal{L}_2$) aligning less as well.}
\label{table: indirect_anchor_exp}
\end{table}

\section{Computational Resources}
For each pretrained mOthelloGPT, we train it with 1 A40 GPU for 24 hours. 

\end{document}